\documentclass[letterpaper, 10 pt, conference]{ieeeconf}  

\IEEEoverridecommandlockouts                              

\usepackage{graphics} 
\usepackage{epsfig} 
\usepackage{mathptmx} 
\usepackage{times} 

\usepackage{xcolor}

\newcommand{\argmax}{\operatornamewithlimits{argmax}}

\newcommand{\E}{\mathbb{E}}

\newcommand{\sinc}{sinc}

\usepackage{relsize}
\usepackage{enumerate}   

\usepackage{amsmath,amsthm,amssymb,amsfonts,dsfont} 
\usepackage{mathtools,thmtools}
\usepackage{algorithm,algorithmicx,listings} 
\usepackage[noend]{algpseudocode} 
\usepackage{graphicx}
\usepackage{subfig}
\usepackage[font={small}]{caption}
\usepackage{comment}

\title{\LARGE \bf
Learning Q-network for Active Information Acquisition
}

\author{Heejin Jeong$^{1}$, Brent Schlotfeldt$^{1}$, Hamed Hassani$^{1}$, Manfred Morari$^{1}$, Daniel D. Lee$^{2}$ and George J. Pappas$^{1}$
\thanks{$^{1}$These authors are with the department of Electrical and Systems Engineering, University of Pennsylvania, Philadelphia, PA 19104, USA
        {\tt\small \{heejinj, brentsc, hassani, morari, pappasg\}@seas.upenn.edu}}%
\thanks{$^{2}$Daniel D. Lee is with the department of Electrical and Computer Engineering, Cornell University, Ithaca, NY 14850, USA
        {\tt\small dd146@cornell.edu}}%
         \thanks{This research is partially supported by ARL CRA DCIST W911NF-17-2-0181.}
}

\begin{document}

\maketitle
\thispagestyle{empty}
\pagestyle{empty}

\begin{abstract}
In this paper, we propose a novel Reinforcement Learning approach for solving the Active Information Acquisition problem, which requires an agent to choose a sequence of actions in order to acquire information about a process of interest using on-board sensors. 
The classic challenges in the information acquisition problem are the dependence of a planning algorithm on known models and the difficulty of computing information-theoretic cost functions over arbitrary distributions. In contrast, the proposed framework of reinforcement learning does not require any knowledge on models and alleviates the problems during an extended training stage. It results in policies that are efficient to execute online and applicable for real-time control of robotic systems. Furthermore, the state-of-the-art planning methods are typically restricted to short horizons, which may become problematic with local minima. Reinforcement learning naturally handles the issue of planning horizon in information problems as it maximizes a discounted sum of rewards over a long finite or infinite time horizon.
We discuss the potential benefits of the proposed framework and compare the performance of the novel algorithm to an existing information acquisition method for multi-target tracking scenarios.
\end{abstract}

\section{Introduction}
\textit{Active information acquisition} is a challenging problem with diverse applications including robotics tasks \cite{kumar2004, sim, karasev, atanasov2013}. It is a sequential decision making problem where agents are tasked with acquiring information about a certain process of interest (target). The objective function for such problems typically takes on the information-theoretic form such as mutual information and Shannon entropy. The theory of optimal experiment design also studies cost functions based on the trace, determinant, or eigenvalues of the information matrix that describes the current information state. A major challenge in active information acquisition problem is the computation of cost functions, such as mutual information, that are difficult to compute for arbitrary probability distributions. Therefore, many approaches can evaluate only short planning horizons, or take greedy actions that are susceptible to local minima \cite{atanasov14}. 

In this paper, we seek to solve the active information acquisition problem using Reinforcement Learning (RL) methods. In RL, a learning agent seeks an optimal or near-optimal behavior through trial-and-error interaction with a dynamic environment \cite{sutton}. As it does not require a knowledgeable external supervisor, RL has been applied to many interesting sequential decision making problems including numerous applications in robotics \cite{kober, peters, jeong}. The recent successes of RL with deep neural networks have enabled a number of existing RL algorithms to be applied in complex environments, such as Atari games and robotic manipulation \cite{mnih2013, mnih2015, Gu2017}. 

There are several advantages of applying RL to active information acquisition problems. One benefit is that we can avoid over-dependence on system models and the dynamics of the target process can be estimated by various state estimation methods such as particle filters \cite{hoffmann} and learning-based approaches for highly non-linear systems \cite{haarnoja, ondruska, ondruska2017}. This dramatically extends the available problem domains. Another benefit is that an RL policy maximizes a discounted sum of future rewards, and thus, it is able to handle infinite planning horizons.
In contrast, existing planning algorithms require prior knowledge on target models and often need to approximate a cost function online, which requires additional assumptions to make the computation tractable \cite{atanasov14}. RL-based approaches demand an extended training stage, but produce policies that are efficient to execute online, especially as compared to long-horizon planning methods. This is necessary particularly for running a robotics system in real-time.

\textbf{Related Work.} A number of methods for solving information acquisition problems for dynamic targets have been studied under various constraints. The solutions vary in the number of robots, the length of the planning horizon, and the probability distribution being tracked. Efficient search-based planning methods have been applied for models assumed to be linear and Gaussian \cite{atanasov14, brent}, and sampling-based algorithms have been used for more complex problems and longer horizons \cite{hollinger}.
Prior work in data-driven information acquisition uses imitation learning via a clairvoyant oracle \cite{choudhury2017,he}. These are supervised learning methods which train a policy to mimic the provided expert trajectories, and thus, require a large and labeled dataset. This method could be more sample efficient than RL, but it is limited to problems having an access to labeled datasets.

Active object tracking is one of the common tasks in active information acquisition. With the substantial achievement of deep learning in the field of computer vision, many deep learning methods for computer vision have been studied in this area \cite{valmadre, choi}. End-to-end solutions for active object tracking are introduced using ConvNet-LSTM and RL algorithms \cite{luo, zhang}. Their reward function is designed specifically for the object tracking problem, for example, in order to reduce a distance between an object and a learning agent. However, maximizing information obtained for a target does not always require the agent to closely follow the target, especially in the case of multiple targets or if a target is highly dynamic. Moreover, the end-to-end approach requires a large number of training samples. 

\textbf{Contributions.} We highlight the following contributions of this paper:
\begin{itemize}
    \item We propose a general RL framework for the active information acquisition problem formulating it as a Markov Decision Process.
    \item We apply the framework to an active target tracking application and use existing Q-network-based deep RL algorithms to learn an optimal policy.
    \item We compare simulation results in various environments with a search-based information acquisition approach in a target tracking scenario. The results demonstrate that the Q-network-based deep RL algorithms are able to outperform the existing method, while making far less assumptions on the underlying problem.
\end{itemize}
\section{Background}
\subsection{Active Information Acquisition}
Suppose that a robot carrying a sensor follows a \textit{discrete time} dynamic model:
\begin{equation}
    x_{t+1} =f(x_t, u_t) \label{eq:state_dynamics}
\end{equation}
where $x_t$ is a robot state and $u_t$ is a control input at time $t$.
The goal of the robot is to actively track $N$ targets of interest using noisy measurements from its sensor governed by some dynamic models:
\begin{equation}
    y_{i,t+1} = g_i(y_{i,t}; \mu_i) \qquad \text{for }i=1,\cdots,N \label{eq:target_dynamics}
\end{equation}
where $y_{i,t}$ is a state of the $i$th target and we composes each target state as $y_{t} = [y_{1,t}^T, \cdots, y_{N,t}^T]^T$. Each target can have its own control policy $\mu_i$.
We denote the sensor measurement signal about the targets as $z_t=[z_{1,t}^T, \cdots, z_{N,t}^T]^T$ and its observation model as $h(\cdot)$:
\begin{equation}
    z_{i,t} = h(x_t, y_{i,t}) \qquad \text{for }i=1,\cdots,N \label{eq:obs_model}
\end{equation}
The available information to the robot at time $t$ is $\mathcal{I}_0 = z_0$ and $\mathcal{I}_t:=(z_{0:t},u_{0:(t-1)})$ for $t > 0$ where the subscript $t_1:t_2$ represents the set of the corresponding variable from time $t_1$ to time $t_2$ for $t_1\leq t_2$.

\textbf{Problem.} (Active Information Acquisition) \textit{Given an initial robot pose $x_0$, a prior distribution of the target state $y_0$, and a planning horizon $T$, the task of the robot is to choose a sequence of functions, $u_t=\pi(\mathcal{I}_t)$, which maximize the mutual information between the target state $y_{t}$ and the measurement set $z_{1:t}$}:
\begin{equation}
    \max_{\pi} \sum_{t=1}^T \mathrm{I}(y_t; z_{1:t}|x_{1:t})
    \label{eq:mi}
\end{equation}
		\begin{align*}
		    s.t.\quad & x_{t+1} = f(x_t, \pi(\mathcal{I}_t)) & t=0,\cdots, T-1\\
		        & y_{i,t+1} = g(y_{i,t}; \mu_i) & t=0,\cdots, T-1 \\
		        & z_{i,t} = h(x_t, y_{i,t}) & t=0, \cdots, T
		\end{align*}

\subsection{Reinforcement Learning}
RL problems can be formulated in terms of an Markov Decision Processes (MDP) described by the tuple, $M=\langle S, A, P, R,\gamma \rangle$ where $S$ and $A$ are state and action spaces, respectively, $P:S\times A\times S\rightarrow [0,1]$ is the state transition probability kernel, $R:S \times A \times S \rightarrow \mathbb{R}$ is a reward function, and $\gamma \in[0,1)$ is a discount factor. A policy, $\pi$, determines the behavior of the learning agent at each state, and it can be stochastic $\pi(s,a)\in [0,1]$ or deterministic $\pi:S \rightarrow A$ Given $\pi$, the value function is defined as $V^{\pi}(s) = \E_{\pi}[\sum^{\infty}_{t=0}\gamma^t R(s_t,a_t, s_{t+1})|s_0=s]$ for all $s \in S$, which is the expected value of cumulative future rewards starting at a state $s$ and following the policy $\pi$ thereafter. The state-action value, $Q$, function is similarly defined as the value for a state-action pair,  $Q^{\pi}(s,a) = \E_{\pi}[\sum^{\infty}_{t=0}\gamma^t R(s_t,a_t,s_{t+1})|s_0=s,a_0 =a]$ for all $s \in S, a \in A$. The objective of a learning agent in RL is to find an optimal policy $\pi^* = \argmax_{\pi}V^{\pi}$. Finding the optimal values, $V^*(\cdot)$ and $Q^*(\cdot,\cdot)$, requires solving the Bellman optimality equation:
\begin{align}
  Q^*(s,a) =& \E_{s'\sim P(\cdot|s,a)}[R(s,a,s') + \gamma \max_{a'\in A}Q^*(s',a')]\\
  V^*(s) =& \max_{a \in A(s)} Q^*(s,a) \quad \forall s \in S\label{eq:Bellman}
\end{align} where $s'$ is the subsequent state after executing the action $a$ at the state $s$. 

When an MDP is unknown or too complicated, RL is used to find an optimal policy. 
One of the most popular RL algorithms is Q-learning, which updates Q values from a temporal difference error using stochastic ascent. When either or both of the state and action spaces are large or continuous, it is infeasible to represent $Q(\cdot, \cdot)$ for all states and actions in a tabular format. Instead, we can use a function approximator to approximately estimate the $Q$ function, $Q(s,a;\xi)\approx Q(s,a)$. When a neural network is used for the function approximator, $\xi$ corresponds to the neural network weights and biases. Deep Q-network (DQN) is a neural network extension to Q-learning which network outputs a vector of action values $Q(s, \cdot; \xi)$ for a given state $s$. DQN solves the difficulty of applying neural network to Q-learning by mainly introducing the use of an additional target $Q$ network and experience replay \cite{mnih2013}. Double DQN is the extension of Double Q-learning with a neural network which reduces the overestimation of DQN by using two sets of neural network weights \cite{hado}. Assumed Density Filtering Q-learning (ADFQ) is a Bayesian counterpart of Q-learning which updates belief distribution over Q values through online Bayesian update algorithm \cite{adfq}. One of the major advantages of ADFQ is that its update rule for Q values takes a non-greedy update with its uncertainty measures and reduces the instability of Q-learning. It has shown that ADFQ with a neural network outperforms DQN and Double DQN when the number of actions of a problem is large. This may be more appropriate for the active information acquisition problem as it can be highly stochastic and potentially has a large number of actions. 

 \begin{figure}[t!]
      \centering
      \includegraphics[width=0.75\columnwidth]{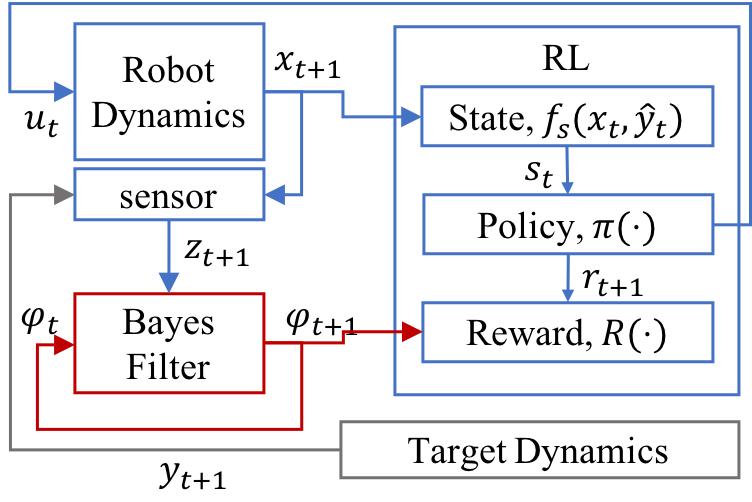}
      \caption{RL framework for Active Information Acquisition}
      \label{fig:diagram}
   \end{figure}
   
\section{RL Framework for Active Information Acquisition}
In order to solve the active information acquisition problem using RL, we first formulate the problem as an MDP. Since the robot does not have access to the ground truth of target states, we could formulate the problem as a Partially Observable Markov Decision Processes (POMDP) that maintains beliefs over states \cite{pomdp}. However, it is known that solving a generic discrete POMDP exactly is wildly intractable \cite{papadim}. Instead, we define a target belief distribution $B(\varphi_i)$ for $i=1,\cdots,N$ where $B(\cdot)$ is a tractable parametric distribution with parameters $\varphi_i$, and $v_{i,t} = \mu_i(y_{i,t})$ is a control input of the target if exists. The belief distribution, or $\varphi_i$, can be updated by a Bayes filter using incoming observations. We explicitly include the belief state as part of the MDP state, and thus, the problem state is expressed as a function of the robot state and the target belief states, $s_t = f_s(x_t, \varphi_{1,t}, \cdots, \varphi_{N, t})$. $f_s(\cdot)$ may vary depending on the application. The action in the MDP is defined as the control input to the robot, $a_t=u_t$. 

The goal of the RL agent in this problem is to find an optimal policy $\pi^*$ that maximizes mutual information (\ref{eq:mi}). Assuming that $y_t$ is independent of the robot path, $x_{1:t}$, the optimization problem now seeks to minimize the differential entropy, $H(y_t|z_{1:t},x_{1:t})$ \cite{atanasov14}. In order to evaluate the entropy resulted after taking an action $a_t$ at the current state $s_t$, a reward is defined by the belief posterior at $t+1$:
\begin{equation}
    R(s_t, a_t, s_{t+1}) \equiv -H(y_{t+1}|z_{1:t+1},x_{1:t+1})
    \label{eq:reward_gen}
\end{equation}
Then, the optimal policy minimizes the discounted cumulative total entropy :
\begin{equation}
    V^{\pi^*}(s) = -\E_{\pi^*}\left[\sum^{\infty}_{t=0}\gamma^t H(y_{t+1}|z_{1:t+1},x_{1:t+1})|s_0=s, \varphi_0 \right]
    \label{eq:value}
\end{equation}
The RL framework for active information acquisition is summarized in Fig.\ref{fig:diagram}. 

\section{Learning Q-network for Active Target Tracking}
In this section, we present a specific RL approach to the active information acquisition problem by focusing on the target tracking application in two-dimensional environment with Gaussian belief distributions. Let the mean and the covariance of the $i$th target belief be $\hat{y}_i$ and $\Sigma_i$, respectively. The RL state, $s$, is defined by the target belief states and the information of surroundings. More formally, 
\[
    s_{i,t} \equiv [\hat{y}_{i,r,t}^{(x)}, \hat{y}_{i,\theta,t}^{(x)}, \dot{\hat{y}}_{i,r,t}^{(x)}, \dot{\hat{y}}_{i,\theta,t}^{(x)}, \log \det \Sigma_{i,t},  \mathbb{I}(y_{i,t} \in \mathcal{O}(x_t))]^T
\]
\[
    s_t \equiv [s_{1,t}^T, \cdots, s_{N,t}^T, o_{r,t}^{(x)}, o_{\theta,t}^{(x)}]^T
\]
where $\hat{y}_{i,r,t}^{(x)}$ is a radial coordinate of the $i$th target belief mean in the robot frame, and $\hat{y}_{i, \theta,t}^{(x)}$ is a polar coordinate of the $i$th target belief mean in the robot frame at time $t$. $\mathcal{O}(x)$ is an observable space from the robot state $x$ and $\mathbb{I}(\cdot)$ is a boolean function which returns 1 if its given statement is true and 0 otherwise. $o_{r,t}^{(x)}$ and $o_{\theta,t}^{(x)}$ are a radial and polar coordinate of the closest obstacle point to the robot in the robot frame, respectively. If there is no obstacle detected, $o_{r,t}^{(x)}$ and $o_{\theta,t}^{(x)}$ are set to its maximum sensor range and $\pi$, respectively In a $SE(3)$ environment, we can use $\hat{y}^{(x)}$ in the spherical coordinate system instead.
We define the action space with a finite number of motion primitives.

Since $p(y_t|z_{1:t}, x_{1:t})$ is a belief posterior, $p(y_t|\varphi_t)$, the differential entropy in (\ref{eq:reward_gen}) is:
\begin{equation}
   H(y_t|z_{1:t},x_{1:t}) = \log \det \Sigma_t + c 
\end{equation}
where $c$ is a constant. Assuming that all target beliefs are independent to each other, $\Sigma$ is a block-diagonal matrix of individual covariances, $\Sigma = diag(\Sigma_1, \cdots, \Sigma_N$), and $\log\det\Sigma = \sum_i \log\det \Sigma_i$. Therefore, we define the reward function in this target tracking problem as:
\begin{align}
    R(s_t, a_t, s_{t+1}) = -&  \kappa_{m} \sum_i \log \det{\Sigma_{i,t+1}} \nonumber \\
    & - \kappa_{d} SD_i[\log \det \Sigma_{i,t+1}] - \kappa_o o_{r,t+1}^{-2}
    \label{eq:reward_tt}
\end{align}
\begin{algorithm}[t!]
\caption{Learning Q-network for Active Target Tracking}
\begin{algorithmic}[1]
\small
\State Randomly initialize a train Q-network, $Q(s,a|\xi)$
\State Initialize a target Q-network, $Q(s,a|\xi')$ with weights $\xi' \leftarrow \xi$
\State Initialize replay buffer
\For{trajectory$=1:M$}
\State Randomly initialize $x_0$, $y_0$, $\hat{y}_0$, $\Sigma_0$
\For{$t=0:T-1$}
\State Execute an action $a_t = \pi^{action}(s_t)$ 
\State Receive the next states $x_{t+1}$ and the measurement $z_{t+1}$
\State $(\hat{y}_{t+1}, \Sigma_{t+1}) \leftarrow$ Bayes Filter$(\hat{y}_{t}, \Sigma_t, z_{t+1})$
\State Compute a reward $r_{t+1} = R(s_t, a_t, s_{t+1})$
\State Update the Q-network
\State Update the state, $s_{t+1}$
\EndFor
\EndFor
\end{algorithmic} \label{table:alg}
\end{algorithm}
The first two terms penalizes the overall uncertainty of the target beliefs and their dispersion (as standard deviation). The dispersion term prevents the robot from tracking only a few targets when not all the targets are within its sensing range at time. The second term discourages the robot to approach toward obstacles or a wall. $\kappa_m$, $\kappa_d$, and $\kappa_o$ are constant factors, and $\kappa_o$ is set to 0 if no obstacle is detected. 

We suggest off-policy temporal difference methods such as DQN, Double DQN, and ADFQ in order to learn an optimal policy for the problem. Although any RL algorithms can be used in this framework, such off-policy temporal difference algorithms are known to be more sample efficient than policy-based RL methods \cite{nachum}. Moreover, an action policy can be different from the update policy in off-policy methods which allow a safe exploration during learning. The algorithm is summarized in Table.\ref{table:alg}. Note that the RL agent does not require any knowledge on the system models (\ref{eq:state_dynamics}), (\ref{eq:target_dynamics}), (\ref{eq:obs_model}) as long as it can observe its state and a reward. Additionally, the RL update is independent from the Bayes filter and it can leverage various state estimation methods. 
\begin{figure}[tb!]
\centering
  \includegraphics[width=0.82\columnwidth]{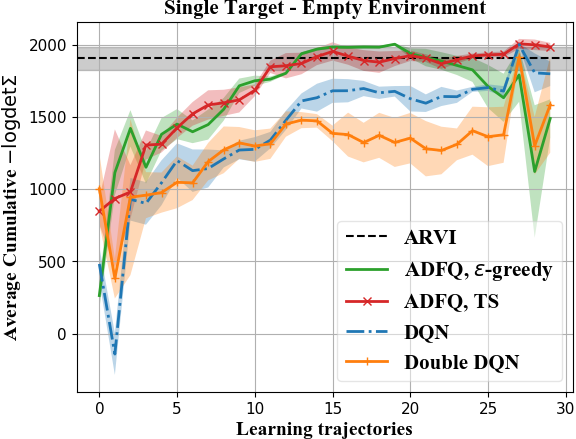}\\
  \vspace{6pt}
  \includegraphics[width=0.82\columnwidth]{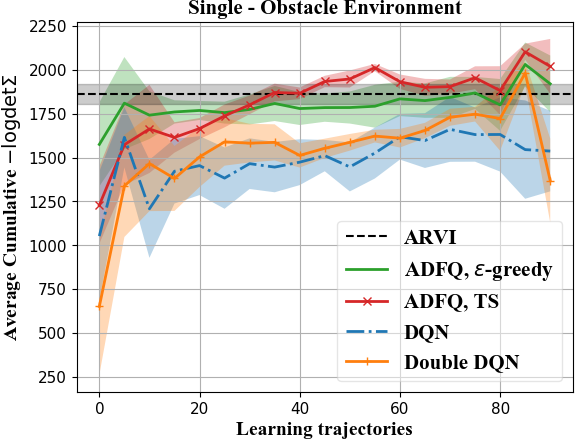}\\
  \vspace{6pt}
  \includegraphics[width=0.82\columnwidth]{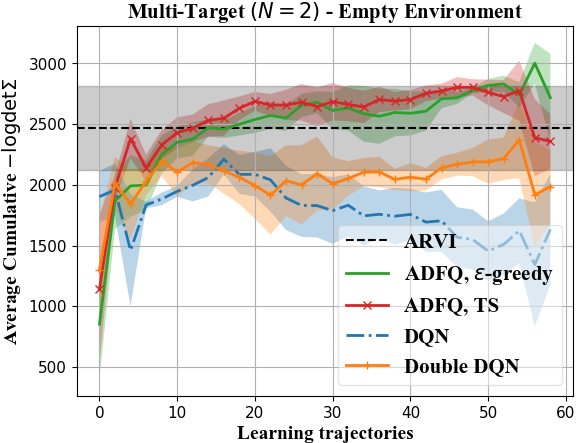}\\
  \vspace{6pt}
  \includegraphics[width=0.82\columnwidth]{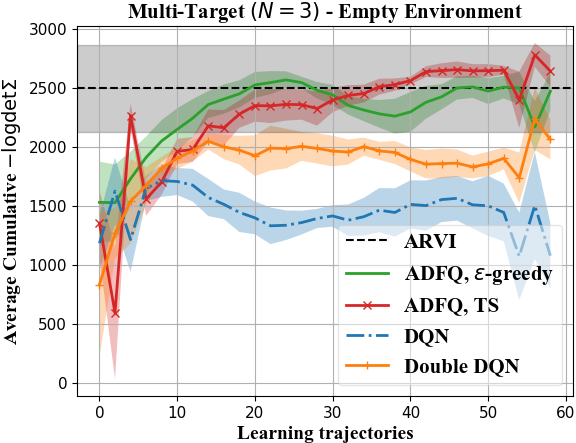}
\caption{\footnotesize Cumulative $-\log \det \Sigma_t$ per trajectory of ADFQ, DQN, and Double DQN during learning compared with ARVI.}
\label{fig:performance}
\end{figure}
\begin{figure*}[tb!]
\centering
\includegraphics[width=\textwidth]{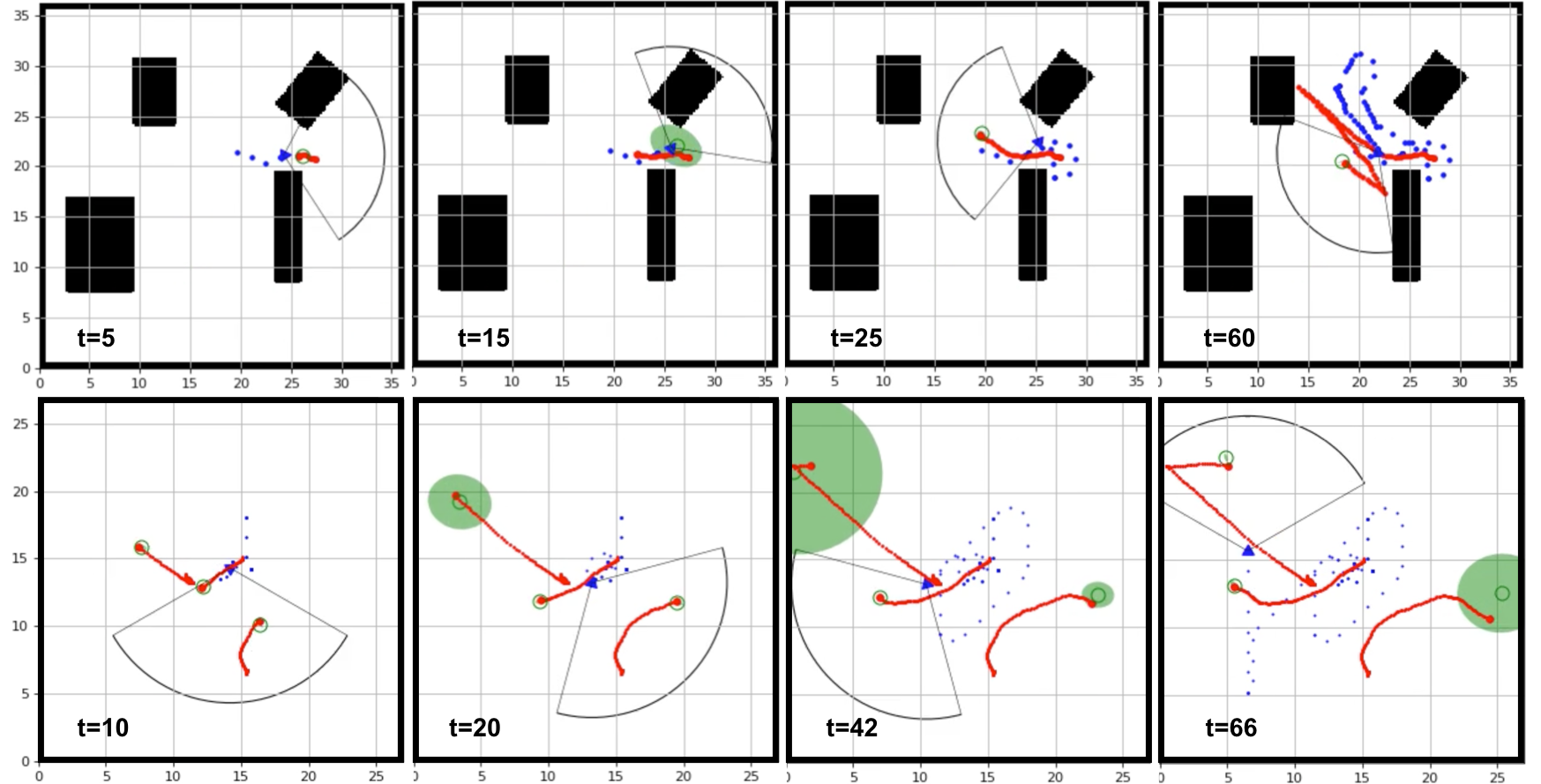}
\caption{Demonstrations of learned ADFQ-TS policies in the obstacle environment with a single target (first row) as well as the empty environment with three target (second row). The time step increases from left to right. Blue triangle: $x_t$, Blue dot: $x_{1:t-1}$, Red dot (big): $y_t$, Red dot (small): $y_{1:t}$, Green circle: $\hat{y}_t$, Green shaded area: $\Sigma_t$.} 
\label{fig:example_obs}
\end{figure*}
\section{Experiments}
To demonstrate the proposed framework, we evaluate it with  ADFQ, DQN, and Double DQN in target tracking problems with different numbers of targets ($N=1,2,3$). $\epsilon$-greedy action policy is used with $\epsilon$ annealed from 1.0 to 0.01 for all algorithms. For ADFQ, we additionally used its Thompson sampling (TS) action policy using its uncertainty estimate for Q-values. 

Furthermore, we compare with the Anytime Reduced Value Iteration (ARVI), an open-source target tracking algorithm, which we use as a baseline. The ARVI uses a linear system model and the Kalman Filter to predict a target trajectory, and then evaluates the mutual information over a search tree with some pruning to ensure finite execution time. The performance of ARVI has been verified in target tracking simulations and real robot experiments in \cite{brent}. The aim is to show the reinforcement learning outperforms this approach, but rather that it achieves a comparable performance while featuring a much more general problem formulation.

The differential drive dynamics of a robot is:
\begin{align}
\begin{bmatrix}
x_{1,t+1} \\ x_{2,t+1} \\ x_{\theta, t+1}
\end{bmatrix} = 
\begin{bmatrix}
x_{1,t} \\x_{2,t} \\ x_{\theta, t}
\end{bmatrix} + 
\begin{bmatrix}
\nu \tau \sinc(\frac{\omega \tau}{2}) \cos (x_{\theta, t} + \frac{\omega \tau}{2})\\
\nu \tau \sinc(\frac{\omega \tau}{2}) \sin (x_{\theta, t} + \frac{\omega \tau}{2})\\
\tau \omega
\end{bmatrix}
\end{align}
where $\tau$ is a sampling period, and $x_{1,t}, x_{2,t}, x_{\theta, t}$ correspond to the elements of $x_t$ in $x$-axis, $y$-axis and polar coordinate at time $t$, respectively. We discretized the action space with pre-defined motion primitives, $A= \{(\nu,\omega) | $ $\nu \in \{0,1,2,3\}$ m/s, $\omega \in \{0,-\pi/2, \pi/2 \}$ rad/s$\}$.  
The objective of the robot is to track the positions and velocities of targets which follows double integrator dynamics with Gaussian noise:
\begin{equation}
    y_{i,t+1} = Ay_{i,t} + w_{i,t}, \qquad w_{i,t} \sim \mathcal{N}(0, W)
    \label{eq:linear_model}
\end{equation}
\[
A=\begin{bmatrix} I_2 &\tau I_2 \\
					0 & I_2  \end{bmatrix}, \qquad W=q \begin{bmatrix} \tau^3/3 I_2 & \tau^2/2 I_2 \\ \tau^2/2 I_2 & \tau I_2 \end{bmatrix}
\]
$q$ is a noise constant factor. When the target is close to a wall or an obstacle, it reflects its direction with a small Gaussian noise. 
We assumed that the target model is known to the robot and updated the target belief distributions using the Kalman Filter. Note that the Kalman Filter can be simply replaced by other Bayes filters or learning-based state estimation methods within the proposed RL framework.

The observation model of the sensor for each target is:
\begin{equation}
    z_{i,t} = h(x_t, y_{i,t}) + v_t, \hspace{3mm} v_t \sim \mathcal{N}\begin{pmatrix}0, V(x_t,y_{i,t}) \end{pmatrix}
\end{equation}
\begin{equation*}
    h(x,y) = \begin{bmatrix}r(x,y)\\\alpha(x,y)\end{bmatrix} := \begin{bmatrix} \sqrt{(y_1 - x_1)^2 + (y_2 -x_2)^2} \\ \tan^{-1} ((y_2 - x_2)(y_1 - x_1)) - x_{\theta} \end{bmatrix}
\end{equation*}
To be used in the Kalman Filter, this model is linearized by computing the Jacobian matrix of $h(y, x)$ with respect to $y$:
{\small \begin{equation*}
\nabla_y h(x,y) = \frac{1}{r(x,y)} 
\begin{bmatrix} (y_1 - x_1) & (y_2 - x_2) & \mathbf{0}_{1x2} \\
-\sin (x_{\theta}+\alpha(x,y)) & \cos (x_{\theta}+\alpha(x,y))& \mathbf{0}_{1x2} \\ \end{bmatrix}
\end{equation*}}
In the experiments, the sensor has a maximum range of 10 meters and its field of view is 120 degree. We assume that the sensor is able to distinguish targets or obstacles. $x_0$ is randomly initialized within the given map and the position components of $y_0$ is also randomly initialized within the maximum offset of 8 meter from the initial robot state. The initial velocity is 0.0. The belief target state follows Gaussian. In order to design the experiment more realistic, the mean position is randomly initialized to have the maximum offset of 5 meter from the target and the covariance, $\Sigma$, is initialized to $30.0 I_4$. We use $\tau=0.5$ and a constant observation noise, $V= \text{diag}(\sigma_r, \sigma_b)$ with $\sigma_r=0.2, \sigma_b=0.01$. 

For the $Q$-network, we used 3 layers with 128 units and a learning rate $0.001$ for a single target, and 3 layers with 256 units and a learning rate $0.0005$ for multiple targets. The target $Q$-network is updated every 50 training steps. The batch size and the replay buffer are 64 and 1000, respectively.

All experiments are obtained with 5 different random seeds for the learning algorithms and 10 random seeds for ARVI. The results are plotted in Fig.\ref{fig:performance}. The darker lines show the mean over seeds and the shaded areas represent standard deviation. The current learned policies from $\xi_t$ were semi-greedily evaluated with $\epsilon=0.05$ for 5 times after trained with a single trajectory (every two trajectories for multi-target experiments). The curves are smoothed by a moving average with window 4.

\subsection{Single Target}
We tested the single target problem in an empty domain $(100\times 100[m^2])$ where there is no obstacle, and therefore, the behavior of the target is more predictable (as there is far less reflection behavior of the target with noise). We also tested a domain with four obstacles as in the first row of Fig.\ref{fig:example_obs}. The noise parameter for the target model is set to $q=0.01$ for both cases and the length of a trajectory is $T=100$ steps. 

The first plot in Fig.\ref{fig:performance} shows that both ADFQ with TS and ADFQ with $\epsilon$-greedy achieved the baseline performance after learning with 13 trajectories. ADFQ-TS showed a more stable performance outperforming the baseline toward the end. Since the belief state mean can quickly diverge from the true state while its covariance is quite small, exploration methods based on state-action uncertainty such as Thompson sampling leads a better performance than $\epsilon$-greedy. 
ADFQ outperformed the baseline in the obstacle environment as well. An example case of a learned policy by ADFQ-TS is presented in the first row of Fig.\ref{fig:example_obs}. As shown, even though it missed the target at $t=15$ and the belief became inaccurate, it quickly adjusted its direction and followed the target keeping it in its range. 

DQN and Double DQN failed to reach the baseline performance in both environments. Although their performances increased with the number of learning trajectories in the empty environment, their performances dramatically dropped in the obstacle environments. This is due to the high stochasticity of the environment as the target changes its path abruptly with noise when it faces an obstacle. 

\subsection{Multi-Target}
We tested the cases of two and three targets in an empty domain $(27\times 27[m^2])$ with $q=0.001$. A longer trajectory, $T=150$, is used in order to evaluate cases where targets diverge and a robot has to keep traveling to minimize the covariances. In both $N=2$ and $N=3$, ADFQ algorithms outperformed or achieved the baseline performance as shown in Fig.\ref{fig:performance}. Additionally, the baseline showed large variances in its performance in both cases while ADFQ algorithms showed fairly lower variances across the trials. 

The most challenging part of these experiments is when not all targets are observable at time. The results indicate that the RL methods can learn a policy which makes a near-optimal decision on when to keep traveling to track all the targets or when to exploit to close targets. 
The learned policy of ADFQ-TS is demonstrated in Fig.\ref{fig:example_obs}. When the targets are not simultaneously observable but not too far from each other, the robot must choose to visit each target sequentially to maintain its belief distribution for every target.
\section{CONCLUSIONS}
In this paper, we introduced a novel RL framework for the active information acquisition problem and developed a detailed approach for solving the active target tracking problem with a $Q$-network-based RL algorithm. The experimental results demonstrated that the RL-based methods can achieve or sometimes outperform the search-based planning algorithm. 
As an initial approach, we used the Kalman filter with a known linear target model in our experiment. Future work will leverage various existing techniques in Bayesian filtering and state estimation within the framework in order to use nonlinear or unknown target models. Additionally, since ADFQ maintains belief distributions over Q-values, we further intend to extend our approach by propagating target state uncertainty to Q-belief distributions.

\addtolength{\textheight}{-12cm}   






\end{document}